\pdfoutput=1

\documentclass[11pt]{article}

\usepackage[]{acl}

\usepackage{times}
\usepackage{latexsym}

\usepackage{graphicx}
\usepackage{amsfonts}
\usepackage{amsmath}

\usepackage{booktabs}
\usepackage[T1]{fontenc}

\usepackage[utf8]{inputenc}

\usepackage{multirow}
\usepackage{microtype}

\title{Generative Biomedical Entity Linking via \\
Knowledge Base-Guided Pre-training and Synonyms-Aware Fine-tuning}

\newcommand*\samethanks[1][\value{footnote}]{\footnotemark[#1]}

 \author{
Hongyi Yuan \thanks{$\quad$Contributed equally.} \space\space
Zheng Yuan \samethanks \space\space\space
Sheng Yu \thanks{$\quad$Corresponded author.} \\
Tsinghua University \\
\texttt{\{yuanhy20,yuanz17\}@mails.tsinghua.edu.cn} \\
\texttt{syu@tsinghua.edu.cn}
}

\date{}

\begin{document}
\maketitle
\begin{abstract}
Entities lie in the heart of biomedical natural language understanding, and the biomedical entity linking (EL) task remains challenging due to the fine-grained and diversiform concept names.
Generative methods achieve remarkable performances in general domain EL with less memory usage while requiring expensive pre-training.
Previous biomedical EL methods leverage synonyms from knowledge bases (KB) which is not trivial to inject into a generative method.
In this work, we use a generative approach to model biomedical EL and propose to inject synonyms knowledge in it.
We propose KB-guided pre-training by constructing synthetic samples with synonyms and definitions from KB and require the model to recover concept names.
We also propose synonyms-aware fine-tuning to select concept names for training, and propose decoder prompt and multi-synonyms constrained prefix tree for inference.
Our method achieves state-of-the-art results on several biomedical EL tasks without candidate selection which displays the effectiveness of proposed pre-training and fine-tuning strategies. The source code is available at \href{https://github.com/Yuanhy1997/GenBioEL}{Github.com/Yuanhy1997/GenBioEL}.

\end{abstract}

\section{Introduction}

Biomedical entity linking (EL) refers to mapping a biomedical mention (i.e., entity) in free texts to its concept in a biomedical knowledge base (KB) (e.g., UMLS \cite{Bodenreider2004}). 
This is one of the most concerned tasks of research in medical natural language processing which is highly related to high-throughput phenotyping \cite{yu2015toward}, relation extraction \cite{li2016biocreative}, and automatic diagnosis \cite{yuan2021efficient}. 

Recent methods in biomedical EL mainly used neural networks to encode mentions and each concept name into the same dense space, then linked mentions to corresponding concepts depending on embedding similarities \cite{sung2020biomedical,rescnn,dualencoder,biocom,arboel}.
Synonyms knowledge has been injected into these similarity-based methods by contrastive learning \cite{sapbert,coder}.
For example in UMLS, concept \textbf{C0085435} has synonyms: \textit{Reiter syndrome}, \textit{Reactive arthritis} and \textit{ReA} which help models to learn different names of a concept entity. 
However, similarity-based methods requires large memory footprints to store representation for each concept which are hard to deploy. 



In general domain, GENRE \cite{genre} 
viewed EL as a seq2seq task which inputs mentions with contexts and outputs concept names token-by-token.  
Mentions, contexts and concepts can be mingled due to the power of transformers, and the model does not need to store representations for each concept during inference.
GENRE pre-trained on Wikipedia EL datasets to boost performances.
However, directly implementing GENRE on biomedical EL cannot harvest satisfying results. 
The gap occurs in two aspects: 
(1) There are no such large-scale human-labeled biomedical EL datasets for pre-training. 
(2) Biomedical concepts may have multiple synonyms. 
We find the results are sensitive to the synonyms selection for training, and simply using a 1-to-1 mapping between names and concepts as \citet{genre,mgenre} may hurt performances.



\begin{figure*}[t]
    \centering
    \includegraphics[width=2.0\columnwidth]{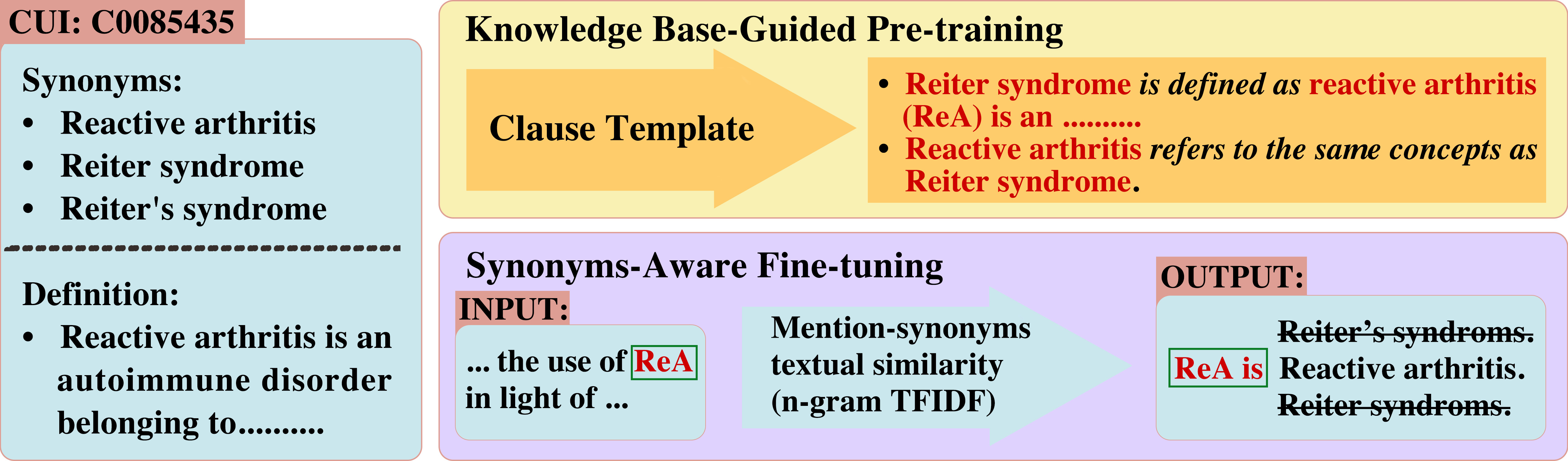}
    \caption{Overview of our approach.}
    \small
    \label{fig:overview}
\end{figure*}

To address the above issues, we propose KB-guided pre-training and synonyms-aware fine-tuning to improve generative EL.
For \textbf{pre-training},
we construct pre-training samples using synonyms and definitions collected from KBs and sentence templates.
KB-guided pre-training has the same format as seq2seq EL, which fills the gap of loss of pre-training corpus.
Compared to the method introduced in \citet{genre}, ours performs better in biomedical EL with fewer resources.
For \textbf{fine-tuning}, we 
propose decoder prompts to highlight mentions.
We find the model tends to generate textually similar names to mentions. Hence textual similar criterion is proposed for selecting concept names during fine-tuning.
During inference, we propose a multi-synonyms constrained prefix tree, which results in significantly improved performance. The overview of our approach is illustrated in Figure~\ref{fig:overview}.

We conduct experiments on various biomedical EL datasets and achieve SOTA results on COMETA, BC5CDR, and AskAPatient (AAP) even without candidate selection. Extensive studies show the effectiveness of our proposed pre-training and fine-tuning schemes.

\section{Approach}
\label{approach}

Define a set of concepts $\mathcal{E}$ as target concepts (i.e. concepts from target KBs). 
For each concept $e\in\mathcal{E}$, we have a set of synonyms names $f(e) = \{s_{e}^i | i\in \{1,\dots, n_e\}\}$.
All the synonyms forms a name set $\mathcal{S}=\bigcup_{e\in\mathcal{E}}\{s_{e}^i | i\in \{1,\dots, n_e\}\}$. 
Names-to-concept mappings can be defined by:$\sigma(s^i_e)=e$ where $\sigma=f^{-1}$.
For mention $m$ with left and right contexts $c_l$ and $c_r$ which gold label is $e_m\in\mathcal{E}$, we need to find the target concept $\hat{e}_m\in\mathcal{E}$. $m$, $c$ and $s$ comprise a sequence of tokens.
\subsection{Seq2seq EL}
\label{seq2seq}
Our model applies an encoder-decoder transformer architecture following \citet{genre}. 
The encoder input is:
$
\texttt{[BOS]} c_{\text{l}} \texttt{[ST]} m \texttt{[ET]} c_{\text{r}} \texttt{[EOS]},
$
where \texttt{[ST]} and \texttt{[ET]} are the special tokens marking $m$\footnote{We use words \textbf{Start} and \textbf{End} as \texttt{[ST]} and \texttt{[ET]} respectively.}.
For the decoder side, unlike GENRE decoding target names directly, we use simple prefix prompts $P_m=$<$m \texttt{ is}$> to strengthen the interaction between mentions and make the decoder side output resemble a natural language sentence:
$
\texttt{[BOS] } m \texttt{ is   } s,
$
where $s$ is a target name belong to label concept $e$.
The training objective of Seq2Seq EL is to maximize the likelihood:
\begin{align*}
p_\theta(s|P_m, c, m) = \prod_{i=1}^{N_s} p_\theta(y_i|y_{<i}, P_m, c, m),
\end{align*}
where $N_s$ is the number of tokens of $s$ and $y_i$ indicates the $i$th token.
The inference of Seq2seq EL applies beam search \cite{sutskever2014sequence} with targets constrained to the name set $\mathcal{S}$ by a prefix tree (constructed by name set $\mathcal{S}$).
Unlike mGENRE \citet{mgenre} using provided candidates to decrease the size of the prefix tree, we use the whole name set $\mathcal{S}$ instead.

\subsection{KB-Guided Pre-training}

As the training data of EL is tiny compared to the vast number of concepts in KB, thus it makes EL for some mentions zero-shot problems.
We want to leverage synonyms knowledge from KB to enhance the model's performance.
Injecting synonyms knowledge to the encoder-only models can be done by contrastive learning \cite{sapbert, coder}.
However, such a paradigm cannot directly apply to encoder-decoder architecture 
as entities are not represented by dense embeddings. 
To mitigate this problem, we construct a pre-training task that shares a similar form as Section~\ref{seq2seq}.
We manually define a set of clause templates to splice with synonyms and definitions in KB to form input synthetic language discourses.
Concretely, we select two synonyms $s_e^a$ and $s_e^b$ and definition $c_e$ of a concept $e\in\mathcal{E}$. 
Then we randomly pick a template to concatenate them to form the encoder input, here we give two examples:
$$
\begin{aligned}
\texttt{[BOS]} & \texttt{[ST]} s_e^a \texttt{[ET]} \textbf{ is defined as } c_e  \texttt{[EOS]} \\
\texttt{[BOS]} & c_e \textbf{ describes } \texttt{[ST]} s_e^a \texttt{[ET]}  \texttt{[EOS]}
\end{aligned}
$$
For the decoder:
$
\texttt{[BOS] } s_e^a \texttt{ is   } s_e^b.
$
$c_e$ is the simulated context and $s_e^b$ is for model to predict. 
If definitions are absent in KB, we will use other synonyms to construct $c_e$.
All templates we used can be found in Appendix~\ref{app:temp}. 

\subsection{Synonyms-Aware Fine-tuning}
\label{ft}
We propose and validate by experiments in Section~\ref{syn} that seq2seq EL is profoundly influenced by the textual similarity between mentions and concept names.
It tends to generate textually similar names.
We select the target name by calculating the character 3-gram TF-IDF similarity \cite{scispacy} between mention $m$ and all synonyms $\{s_e^i\}$ and choosing the most similar one as 
\[
s=\arg\max_{s\in \{s_e^i\}}\cos({\rm TFIDF}(m), {\rm TFIDF}(s)).
\]
By the textual similarity criterion, we manually reduce the difficulty of fine-tuning.
We do not use this criterion for pre-training since we want it to learn various synonyms to improve generalization.

Different from GENRE using only one canonical name for each concept for prefix tree, 
we use multi-synonym names (i.e. $\mathcal{S}$) to construct prefix tree.
During inference, we apply prefix tree constrained beam search to decode the name $\hat{s}_m$, and map to the concept $\hat{e}_m=\sigma(\hat{s}_m)$ via N-to-1 names-to-concept mapping $\sigma$. 


\section{Experiments}


\subsection{Datasets and KBs}


\begin{table}
\small
\centering
\begin{tabular}{lc}
\hline 
\textbf{Model} & BC5CDR \\
\hline
\citet{dualencoder}  &  84.8  \\
\citet{clustering} &  91.3   \\
\citet{dataintegration}  & $91.9_{\pm 0.2}$    \\
\hline
FT (Ours) & $92.6_{\pm 0.1}$\\
PT + FT (Ours) & $\textbf{93.3}_{\pm 0.2}$ \\
\hline
\end{tabular}
\caption{Recall$@1$ on BC5CDR test dataset. \citet{sung2020biomedical,sapbert,rescnn} evaluate it by splitting into two subsets which is a easier setting and we do not compare to them. FT corresponds to fine-tuning and PT corresponds to pre-training.}
\label{tab:bc5}
\small
\end{table}

\paragraph{Pre-training} We use a subset of UMLS \textit{st21pv} \cite{medmentions} as $\mathcal{E}$ for pre-training. 
It contains 2.37M concepts, where 160K concepts contain definitions and 1.11M concepts have multiple synonyms.
While pre-training, we iterate concepts and synonyms to construct inputs and outputs.

\paragraph{Fine-tuning} We evaluate our model on BC5CDR \cite{li2016biocreative}, NCBI \cite{dougan2014ncbi}, COMETA \cite{cometa} and AAP \cite{limsopatham-collier-2016-normalising}.
These benchmarks focus on different entity types, including disease, chemicals, and colloquial terms.
For fair comparison, we follow \citet{dataintegration} in BC5CDR, \citet{rescnn} in NCBI and COMETA, and \citet{limsopatham-collier-2016-normalising} in AAP to construct dataset splits and target KB concepts $\mathcal{E}$.
Name set $\mathcal{S}$ is constructed by synonyms from UMLS and original KB which is detailed in Appendix~\ref{app:dedup}.
Datasets summaries are shown in Appendix~\ref{app:dataset}.
We use recall$@1$/$@5$ as metrics for performance illustration, and recall$@5$ results are listed in Appendix~\ref{app:add}.

\subsection{Implementation Details}

We use \texttt{BART-large} \cite{lewis-etal-2020-bart} as the model backbone.
We pre-train and fine-tune our model using teacher forcing \cite{williams1989learning} and label smoothing \cite{szegedy2016rethinking} which are standard in seq2seq training.
The hyperparameters can be found in Appendix~\ref{app:hyp}.




\subsection{Main Results}

Table~\ref{tab:bc5} and~\ref{tab:main} compare the recall$@1$ with state-of-the-art(SOTA) methods.
Our method with KB-guided pre-training exceeds the previous SOTA on BC5CDR, COMETA and AAP, which is up to 1.4 on BC5CDR, 1.3 on COMETA and 0.3 on AAP. 
Our method also shows superiority over previous SOTA on BC5CDR and COMETA without pre-training.
Besides, our method shows competitive results on NCBI compared to SOTA, and we further analyse the results of NCBI through case studies in Appendix~\ref{app:case}.

\begin{table}
\small 
\centering
\begin{tabular}{lccc}
\hline 
\textbf{Model} & NCBI & COMETA & AAP \\
\hline
\citet{sung2020biomedical}   & 91.1  & 71.3 & 82.6   \\
\citet{sapbert}     & 92.3  & 75.1 & 89.0  \\
\citet{rescnn}    &  92.4  & 80.1&-  \\ 
\hline
FT (Ours)  & $91.6_{\pm 0.1}$ & $80.7_{\pm 0.2}$ & $88.8_{\pm 0.1}$   \\
PT + FT (Ours)  & $\textbf{91.9}_{\pm 0.2}$ & $\textbf{81.4}_{\pm 0.1}$ & $\textbf{89.3}_{\pm 0.1}$ \\
\hline
\end{tabular}
\caption{Recall$@1$ on NCBI-disease, COMETA and AskAPatient test datasets.}
\label{tab:main}
\small 
\end{table}


\section{Discussion}


\paragraph{Does pre-training help?}

On all datasets, KB-guided pre-training improves fine-tuning consistently, which is 0.7 on BC5CDR, 0.3 on NCBI, 0.7 on COMETA, and 0.5 on AAP.
To better understand KB-guided pre-training, we conduct ablation studies.
We compare different pre-trained models without fine-tuning in Table~\ref{tab:pre-train}.
BART fails to link mentions due to the mismatch of the pre-training task.
GENRE has been pre-trained on the large-scale BLINK dataset \cite{wu2019zero}, and it obtains a decent ability to disambiguate biomedical mentions.
Our pre-trained model shows improvement on BC5CDR and COMETA compared to GENRE with fewer pre-training resources (6 GPU days vs. 32 GPU days).
We then conduct synonyms-aware fine-tuning on different pre-trained models in Table~\ref{tab:pre-train}. 
Our pre-trained model outperforms BART (+0.8 on BC5CDR, +0.5 on COMETA) and GENRE (+0.4 on BC5CDR, +0.6 on COMETA) which proves the effectiveness of pre-training.

\begin{table}[t]
\small
\centering
\begin{tabular}{lcc}
\hline 
Init & BC5CDR & COMETA \\
\hline 
\bf{Not Fine-tune} \\
BART &6.2 &4.1\\
GENRE &38.3 & 23.8\\
KB-guided & 42.4&33.1 \\
\hline 
\multicolumn{3}{l}{\bf{Synonyms-aware Fine-tune}} \\
BART & 92.5 & 80.9 \\
GENRE & 92.9 & 80.8\\
KB-guided & \textbf{93.3} & \textbf{81.4} \\
\hline 
\end{tabular}
\caption{Recall$@1$ for BC5CDR and COMETA test dataset using different initial checkpoints. We only use left context for BART in decoding and omit decoding prompts for GENRE when not fine-tuning which are consistent with their pre-training.}
\label{tab:pre-train}
\small
\end{table}

\paragraph{Selection of Names}
\label{syn}

We ablate the selection of target names $s$ for fine-tuning: (a) proposed TF-IDF similarity; (b) the shortest name in a concept; (c) randomly sampled name in a concept.
We also compare how to construct $\mathcal{S}$ for inference: (i) using all synonyms from UMLS and target KB; (ii) using the shortest name for a concept; (iii) using randomly sampled name for a concept. 
We note (i) establish an N-to-1 mapping from synonym name to concept, while (ii) and (iii) establish a 1-to-1 mapping.
We conduct experiments on available combinations.
From Table~\ref{tab:syn}, we conclude that (1) N-to-1 mapping performs better than 1-to-1 mappings during inference, which means synonyms can boost performances.
Using one synonym like GENRE degrades performances.
(2) Textual similarity criterion performs better than shortest or sampled names when training. 

\begin{figure}[t]
    \centering
    \includegraphics[width=0.9\columnwidth]{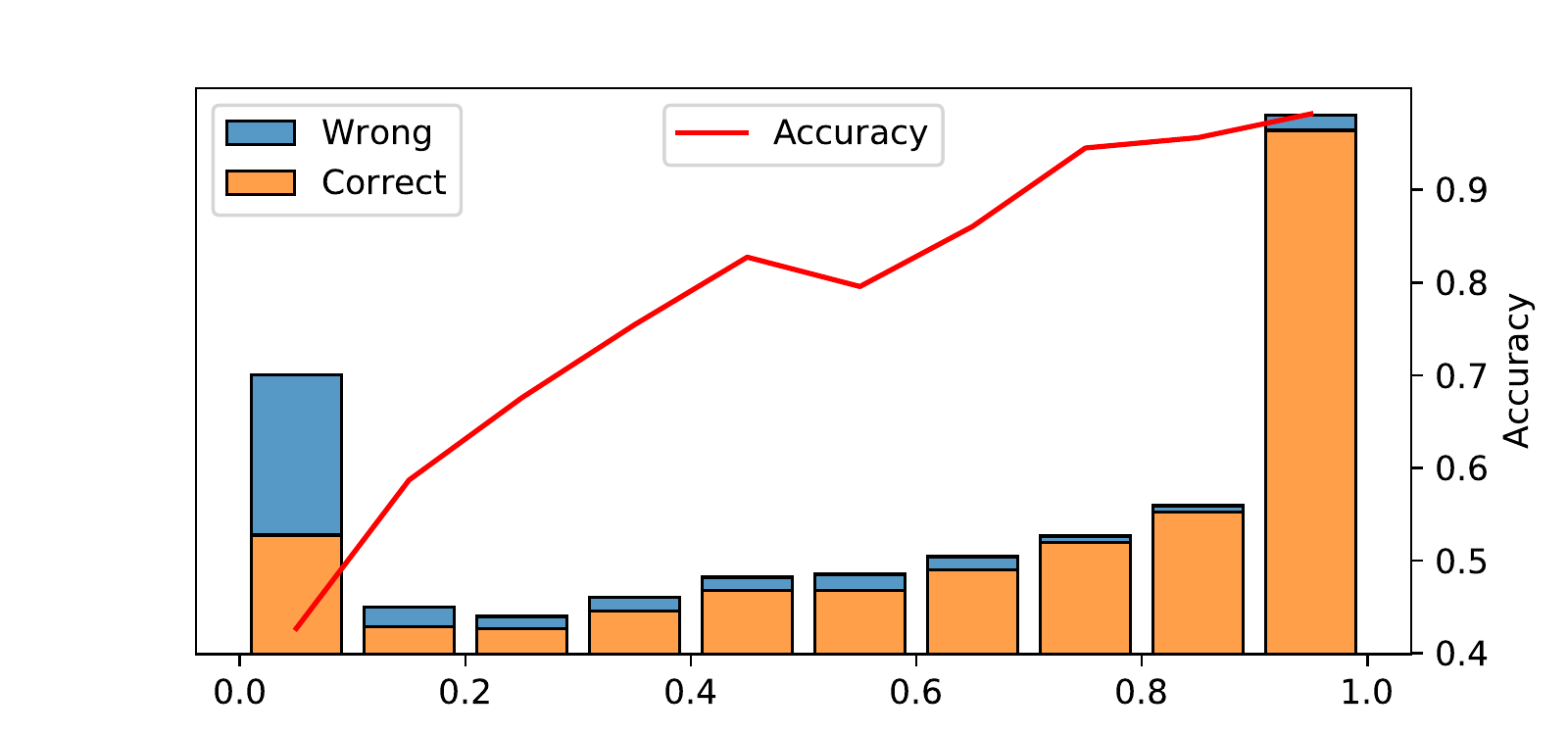}
    \caption{The accuracy of our model on COMETA dataset. The X-axis represents the TF-IDF similarities between mention $m$ and the selected target names $s$.}
    \label{fig:tfidf}
\end{figure}

We also check the accuracies of different TF-IDF similarity sample groups on COMETA trained with using all synonyms as $\mathcal{S}$ and TF-IDF for selecting target names $s$.
From Figure~\ref{fig:tfidf}, we find the distribution of TF-IDF similarity between mentions and selected names is polarized, and the accuracy increases along with textual similarities which prove textually similar targets are easy to generate. This phenomenon validates the advantage of selecting textually similar names for fine-tuning.

\begin{table}
\small
\centering
\begin{tabular}{llccc}
\hline 
$s$ & $\mathcal{S}$ & Prompt & BC5CDR & COMETA \\
\hline 
TF-IDF & All & $\surd$ & \textbf{93.3} & \textbf{81.4} \\
Shortest & All &$\surd$ &87.2 &80.6 \\
Sample & All &$\surd$ &86.9 & 80.8\\
Shortest & Shortest &$\surd$ & 76.3 & 77.5 \\
Sample & Sample &$\surd$ & 72.4 & 77.8 \\
\hline 
TF-IDF & All & $\times$ & 93.0 & 80.9 \\
\hline
\end{tabular}
\caption{Recall$@1$ for BC5CDR and COMETA test dataset using different $s$ for training, different $S$ for inference and applying decoder prompts or not.}
\label{tab:syn}
\small
\end{table}

\paragraph{Decoder Prompting}

One difference between GENRE and ours is using prompt tokens on the decoder side.
Prompting has shown improvement on various NLP tasks \cite{liu2021pre}. 
Here, prompt tokens serve as informative hints by providing additional decoder attention queries and making the outputs resemble language models' pre-training tasks. 
We test dropping the prompt tokens, and Table~\ref{tab:syn} shows degraded performances (-0.3 on BC5CDR and -0.5 on COMETA). The results illustrate the improvement brought by decoder prompting.

\paragraph{Sub-population Analysis}
We list several sub-populations of the BC5CDR benchmark to illustrate the model's performance on different fine-grained categories of mentions. The details of sub-populations are shown in Appendix~\ref{app:subpop}

\begin{table*}
\small
\centering
\begin{tabular}{lccccc}
\hline 
 Subset& \multicolumn{2}{c}{\citet{dataintegration}}  & \multicolumn{2}{c}{Ours} & Sample Size \\
 &Baseline&Full & FT & PT + FT \\
\hline
Overall &89.5 &91.3&\underline{92.5} &\textbf{93.4}&9.65k \\
Single Word Mentions &91.4 &92.9&\underline{95.8} & \textbf{96.6}&7.04k\\
Multi-Word Mentions&84.5 &\textbf{86.8} &\underline{85.3} &84.8&2.62k\\
Unseen Mentions &75.3 &79.6&\underline{81.8} &\textbf{83.3}&3.28k\\
Unseen Concepts &69.7 &77.5& \underline{86.3}&\textbf{86.9}&2.16k\\
Not Direct Match &\underline{89.4} &\textbf{91.9}&83.0& 84.3&3.83k\\
Top 100 &97.3&97.2& \underline{97.9}&\textbf{98.1}&3.31k\\
\hline
\end{tabular}
\caption{Accuracy over sub-populations on BC5CDR of our proposed methods and \citet{dataintegration}. The best results are presented in bold letters and the second best results are highlighted by underlines.}
\label{subpopresult}
\end{table*}

Our model's performance on different sub-populations of BC5CDR is shown in Table \ref{subpopresult}. Through the results, we have several findings:
\begin{enumerate}
    \item Compared with \citet{dataintegration}, our method shows superiority over most of the sub-populations of BC5CDR. Our method without pre-training outperforms the data-augmented version of \citet{dataintegration}.
    \item Our method surpasses \citet{dataintegration} by the largest margin on \textbf{Unseen Concepts}. One possible explanation is that our generative method learns the linkage between mentions and contextual information better, thus gaining superior zero-shot performance.
    \item Our KB-guided pre-training gains improvement universally on most subsets. This reflects the effectiveness of the pre-training task.
    \item Popular concepts and single-word mentions are more easily resolved compared to unseen mentions or concepts and multi-word mentions, respectively. Unsurprisingly, mentions with less training resources and longer textual forms are more difficult. The  lengths of mentions may challenge the generative model.
\end{enumerate}




\section{Related Work}

\paragraph{Biomedical EL} is an important task in biomedical NLP.
Classification-based methods used a softmax layer for classification \cite{limsopatham-collier-2016-normalising,miftahutdinov2019deep} which consider concepts as category factors and lost information of concept names.
Recent methods \cite{sung2020biomedical,sapbert,rescnn,dualencoder,biocom,arboel,coder} encoded mentions and names into a common space and disambiguated mentions by nearest neighbors. 
\citet{clustering} and \citet{dataintegration} adopted a retrieve-and-rerank framework to boost performances. 
\citet{dataintegration} emphasized the lack of training samples in EL and augmented data using Wikipedia and PubMed, while our pre-training corpus constructed by KB and templates can serve as good supplementary training data.


\paragraph{Generative EL}
\citet{genre} proposed to view EL as a seq2seq problem that got rid of hard negative sampling during training and required less memory at inference. 
The shortage is it demanded vast training sources (11 GB training data and 32 GPU days) to achieve competitive performance. 
\citet{mgenre} explored dealing synonyms in multilingual generative EL by adding language identifiers that cannot be directly implemented in biomedical EL.

\section{Conclusion}

To the best of our knowledge, our work is the first to explore generative EL in the biomedical domain. 
We inject synonyms and definition knowledge into the generative language model by KG-guided pre-training. We emphasize the synonym selection issue and propose synonyms-aware fine-tuning by considering the textual similarity. Decoding prompts are also introduced to improve the model's performance. Our model sets new state-of-the-art on different biomedical EL benchmarks.
GENRE shows that well-selected candidate sets can improve seq2seq EL, and we believe this will further boost our performances.

\section*{Acknowledgements}

We thank Shengxuan Luo and Chuanqi Tan for fruitful discussions and advises, and Xixi Mo for drawing the overview figure. We appreciate the anonymous reviewers for their helpful comments and suggestions. This work was supported by the National Natural Science Foundation of China (Grant No. 12171270), the Natural Science Foundation of Beijing Municipality (Grant No. Z190024), and the International Digital Economy Academy.

\bibliographystyle{acl_natbib}
\bibliography{acl2020}

\appendix









\section{Dataset Summary and Statistics}
We pre-process the datasets by the following procedures: 
(1) the abbreviations in the texts are expanded using AB3P \cite{ab3p}; 
(2) the texts are lower-cased, and the beginning and ending of a mention are marked by two special tokens \texttt{[ST]} and \texttt{[ET]}; 
(3) the overlapping mentions and mentions absent from the target KB are discarded.

\label{app:dataset}

\paragraph{BC5CDR} \cite{li2016biocreative} is a benchmark for biomedical entity recognition and disambiguation. The dataset annotates 1500 PubMed article abstracts with 4409 chemicals, 5818 diseases entities, and 3116 chemical-disease interactions. All the annotated entities are mapped to MeSH ontology, which is a smaller medical vocabulary that comprises a subset of UMLS \cite{Bodenreider2004}. In this work, in consideration of fairness, we follow two most recent works \cite{clustering,dataintegration} that use MeSH contained in UMLS 2017 AA release to construct the target knowledge base. 

\paragraph{NCBI} \cite{dougan2014ncbi} contains a corpus of 793 PubMed abstracts. It consists of 6892 annotated disease mentions of 790 unique disease concepts. All the mentions are labeled with concepts in MEDIC ontology \cite{davis2012medic}. MEDIC is a medical dictionary that merges the diseases concepts, synonyms, and definitions in MeSH and OMIM and is composed of 9700 unique diseases. In our work, we used the processed data and the target ontology provided by BioSyn \cite{sung2020biomedical} and ResCNN \cite{rescnn}. We followed their works to construct our training, developing, and testing data. 


\paragraph{COMETA} \cite{cometa} consists of 20k English biomedical entity mentions from publicly available and anonymous health discussions on Reddit. All the mentions are expert-annotated with concepts from SNOMED CT. We use the “stratified (general)” split and follow the evaluation protocol of SapBert \cite{sapbert} and ResCNN \cite{rescnn}.

\paragraph{AskAPatient} \cite{limsopatham-collier-2016-normalising} is a dataset containing 8,662 phrases of social media language. Each phrase can be mapped to one of the 1,036 medical concepts from SNOMED-CT and AMT (the Australian Medicines Terminology). The samples in AskAPatient do not include contextual information and mentions can only be disambiguated by phrases per se. We follow the experimental settings from works of \citet{sung2020biomedical} and \citet{limsopatham-collier-2016-normalising} and apply the 10-fold evaluation protocol.

Statistics of above-mentioned datasets are listed in Table~\ref{tab:stat}.

\begin{table*}
\small
\centering
\begin{tabular}{lcccc}
\hline 
 & NCBI & BC5CDR & COMETA & AskAPatient\\
\hline
Target names $\|\mathcal{S}\|$ &108,071 &809,929 &904,798&3,381 \\
Target concepts$\|\mathcal{E}\|$ &14,967 &268,162 & 350,830&1,036 \\
Train samples&5784 &9,285&13,489&16,826 \\
Dev samples&787 &9,515&2,176&1,663 \\
Test samples& 960 &9,654&4,350&1,712 \\
\hline
\end{tabular}
\caption{Basic statistics of NCBI-disease, BC5CDR, COMETA, AskAPatient datasets and the corresponding target knowledge bases.}
\label{tab:stat}
\end{table*}

\section{License and Availability of Resources}

BC5CDR, NCBI, COMETA, and AskAPatients are all publicly available datasets on the Internet. Their target KBs MeSH, MEDIC, and SNOMED CT are covered by UMLS Metathesaurus License. One can require such a license by signing up for a UMLS terminology services account to access KBs mentioned above.

\section{Additional Experiment Results}
\label{app:add}

\subsection{Recall@5 Results}

We show the Recall@5 result of our method with and without pre-training in Table \ref{tab:recall5}. 
\begin{table}
\small
\centering
\begin{tabular}{lcccc}
\hline 
Dataset & NCBI & BC5CDR & COMETA & AAP\\
\hline
FT &$95.6_{\pm 0.1}$&$95.3_{\pm 0.1}$&$88.7_{\pm 0.2}$&$95.6_{\pm 0.1}$\\
PT+FT&$96.3_{\pm 0.1}$&$95.8_{\pm 0.2}$&$88.2_{\pm 0.1}$&$96.0_{\pm 0.1}$\\
\hline
\end{tabular}
\caption{The Recall@5 results of our proposed method on different benchmarks.}
\label{tab:recall5}
\end{table}

\subsection{Sub-population Analysis}
\label{app:subpop}

Following \citet{dataintegration}, we split test samples into different sub-populations. The details of different sub-population categories we use is shown in Table~\ref{subpopu}.

\begin{table*}
\small
\centering
\begin{tabular}{ll}
\hline 
 Subset& Definition \\
\hline
Overall & Full set of the data\\
Single Word Mentions & Mentions that have one sole word.\\
Multi-Word Mentions& Mentions that have multiple words(Separated by blank spaces).\\
Unseen Mentions & Mentions not existing in fine-tuning set.\\
Unseen Concepts & Concepts not existing in fine-tuning set.\\
Not Direct Match & Mentions that are not a synonym of the target concept in KB.\\
Top 100 & Mentions that mapped to the top 100 concepts in existing frequency in fine-tuning set.\\
\hline
\end{tabular}
\caption{Definitions of sub-populations on BC5CDR.}
\label{subpopu}
\end{table*}

\section{Case Study}
\label{app:case}

We provide case studies on the NCBI-disease benchmark to give an insight and justification of the performance of our method. For the case of mention \textbf{\textit{colorectal adenomas}} which is annotated with \textbf{D018256 adenomatous polyp}, the mention exists for 6 times in the test set (account for 0.63 score of Recall@1). Our model fails to correctly disambiguate all such mentions while linking the mention to other concepts \textbf{D000236 colorectal adenomas} or \textbf{D003123 hereditary nonpolyposis colorectal cancer}. In the training set, the mention \textit{colorectal adenomas} exists for two times and is annotated with \textbf{D003132} and \textbf{D000236} respectively. Thus, through this case, we can see (1) our model learns the information contained in the training set; (2) such inconsistent test samples are hard to  disambiguate correctly. 

\section{Implementation Details}
In this section, we provide more details of our experiments.
\subsection{Knowledge Base Pre-processing}
\label{app:dedup}

Given different knowledge bases, we pre-process their content by the following procedure:
\begin{enumerate}
    \item For each concept, we include all its synonyms from the original target KB. We also expand the synonym set for each concept using UMLS. We use the 2017 AA Active Release of UMLS. 
    \item For synonyms in the expanded name set $\mathcal{S}$, we lowercase them and remove the symbols (e.g., dash line - or comma ,). 
    \item There may exist a name as a synonym for multiple concepts. We de-duplicate these overlapped synonyms by removing the synonym from the concept with more other synonyms to avoid the unbalanced number of synonyms in each concept. 
\end{enumerate}
It is worth noticing that we do not de-duplicate the synonyms in the target KB of NCBI in consideration of comparison fairness.
In the previous works, a mention link to multiple concepts, and correct disambiguation is claimed if the target concept is hit by one of the predicted concepts in NCBI.

\subsection{Pre-training Clause Templates}
We list pre-training clause templates we used in Table~\ref{temtable}. For those concepts containing only 2 synonyms, $s^a_e$ and $s^b_e$ are the two synonyms respectively and $c_e$ is the same as $s^b_e$. For those concepts containing only 1 sole synonym, $s^a_e$, $s^b_e$ and $c_e$ are the same. 
\label{app:temp}
\begin{table*}
\small
\centering
\begin{tabular}{ccll}
\hline 
\multicolumn{2}{c}{Concepts}& \multicolumn{2}{c}{Templates} \\
Definition & >2 synonyms &  Encoder Side & Decoder Side\\
\hline 
\multirow{5}{*}{$\surd$}  & \multirow{5}{*}{$\surd$ / $\times$} &  $s^a_e$ <\texttt{is defined as}> $c_e$.&\multirow{5}{*}{$s^a_e$ is $s^b_e$.} \\
&&  $s^a_e$ <\texttt{is described as}> $c_e$.&\\
&&  $c_e$ <\texttt{are the definitions of}> $s^a_e$.&\\
&&  $c_e$ <\texttt{describe}> $s^a_e$.&\\
&&  $c_e$ <\texttt{define}> $s^a_e$.&\\
\hline
\multirow{4}{*}{$\times$}  & \multirow{4}{*}{$\surd$} &  $c_e$ <\texttt{are the synonyms of}> $s^a_e$.&\multirow{4}{*}{$s^a_e$ is $s^b_e$.}\\
& &  $c_e$ <\texttt{indicate the same concept as}> $s^a_e$.&\\
& &  $s^a_e$ <\texttt{has synonyms such as}> $c_e$.&\\
& &  $s^a_e$ <\texttt{refers to the same concepts as}> $c_e$.&\\
\hline
\multirow{4}{*}{$\times$}  & \multirow{4}{*}{$\times$}&  $c_e$ <\texttt{is}> $s^a_e$.&\multirow{4}{*}{$s^a_e$ is $s^b_e$.}\\
& &  $c_e$ <\texttt{is the same as}> $s^a_e$.&\\
& &  $s^a_e$ <\texttt{is}> $c_e$.&\\
& &  $s^a_e$ <\texttt{is the same as}> $c_e$.&\\
\hline 
\end{tabular}
\caption{The templates used for constructing pre-training samples for different kinds of concepts. $s^a_e$ is the selected synonym as the input mention, $s^b_e$ is the synonym selected as the decoding target. $c_e$ is the contextual information comprised by definition, other synonyms, or mention itself. The template words are between <>, and we omit the special tokens for marking mentions for conciseness.}
\label{temtable}
\end{table*}

\subsection{Experiment Parameters}

Our model contains 406M parameters with 12-layer transformer encoders and 12-layer transformer decoders.
We list the hyper-parameters of our model for KB-guided pre-training and synonyms-aware fine-tuning on different benchmarks in Table \ref{hyper para}.
For pre-training, we heuristically select our parameters.
For fine-tuning, we tune training steps among $\{20000, 30000, 40000\}$ on development set.
For BC5CDR and COMETA, we use learning rate as $1e-5$ and warmup steps as 500.
For AskAPatient and NCBI, we search learning rate among $\{5e-6,8e-7,3e-7\}$, and do not use warmup.
We only evaluate our model at the end of training.
For each benchmark, we run three times to calculate means and standard deviations.
\label{app:hyp}
\begin{table*}[ht]
\centering
\small
\begin{tabular}{lccccc}
\hline
Parameters & Pre-train & BC5CDR & COMETA & AskAPatient & NCBI\\
\hline
Training Steps & 80,000  & 20,000 & 40,000 & 30,000 & 20,000 \\
Learning Rate & 4e-5  & 1e-5 & 1e-5 & 5e-6 & 3e-7 \\
Weight Decay & 0.01 & 0.01 & 0.01 & 0.01& 0.01\\
Batch Size & 384 &8 & 8 & 8 & 8 \\
Adam $\epsilon$ & 1e-8  & 1e-8 & 1e-8 & 1e-8 & 1e-8 \\
Adam $\beta$ & (0.9,0.999)  & (0.9,0.999) & (0.9,0.999) & (0.9,0.999) & (0.9,0.999)\\
Warmup Steps & 1,600  & 500 & 500 & 0 & 0 \\
Attention Dropout & 0.1 & 0.1 & 0.1 & 0.1 & 0.1 \\
Clipping Grad & 0.1  & 0.1 & 0.1 & 0.1 & 0.1 \\
Label Smoothing & 0.1 & 0.1 & 0.1 & 0.1 & 0.1 \\
\hline
\end{tabular}
\caption{Hyper-parameters for KB-guided pre-training and synonyms-aware fine-tuning.
}
\label{hyper para}
\end{table*}

\subsection{Computational Resource}

For our KB-guided pre-training, we implement our model on 6 A100 GPU with 40 GB memory with the help of DeepSpeed ZeRO 2 \cite{zero2} and train for 1 day. For fine-tuning on different benchmarks, we implement our model on 1 A100 GPU.

\end{document}